# Prospects for Theranostics in Neurosurgical Imaging: Empowering Confocal Laser Endomicroscopy Diagnostics via Deep Learning

^Mohammadhassan Izadyyazdanabadi[1, 2], ^Evgenii Belykh[2,3], Michael A. Mooney[2], Jennifer M. Eschbacher[2], Peter Nakaji[2], Yezhou Yang[1] and Mark C. Preul[2]

1 Active Perception Group, School of Computing, Informatics, and Decision Systems Engineering, Arizona State University, Tempe AZ 85281, USA
2 Neurosurgery Research Laboratory, Department of Neurosurgery, Barrow Neurological Institute, St. Joseph's Hospital and Medical Center, Phoenix, AZ 85013, USA
3 Department of Neurosurgery, Irkutsk State Medical University, Irkutsk, 664003, Russia

**Correspondence*:**
**Mark C. Preul, M.D.**
**mark.preul@dignityhealth.org**

^ Authors MI and EB share first authorship

ABSTRACT
Confocal laser endomicroscopy (CLE) is an advanced optical fluorescence imaging technology that has potential to increase intraoperative precision, extend resection, and tailor surgery for malignant invasive brain tumors because of its subcellular dimension resolution. Despite its promising diagnostic potential, interpreting the gray tone fluorescence images can be difficult for untrained users. CLE images can be distorted by motion artifacts, fluorescence signals out of detector dynamic range, or may be obscured by red blood cells, and thus interpreted as nondiagnostic. However, just a single CLE image with a detectable pathognomonic histological tissue signature can suffice for intraoperative diagnosis. Dealing with the abundance of images from CLE is not unlike sifting through a myriad of genes, proteins, or other structural or metabolic markers to find something of commonality or uniqueness in cancer that might indicate a potential treatment scheme or target. In this review we provide a detailed description of bioinformatical analysis methodology of CLE images that begins to assist the neurosurgeon and pathologist to rapidly connect on-the-fly intraoperative imaging, pathology, and surgical observation into a conclusionary system within the concept of theranostics. We present an overview and discuss deep learning models for automatic detection of the diagnostic CLE images and discuss various training regimes and ensemble modeling effect on power of deep learning predictive models. Two major approaches reviewed in this paper include the models that can automatically classify CLE images into diagnostic/nondiagnostic, glioma/nonglioma, tumor/injury/normal categories and models that can localize histological features on the CLE images using weakly supervised methods. We also briefly review advances in the deep learning approaches used for CLE image analysis in other organs. Significant advances in speed and precision of automated diagnostic frame selection would augment the diagnostic potential of CLE, improve operative workflow and integration into brain tumor surgery. Such technology and bioinformatics analytics lend themselves to improved precision, personalization, and theranostics in brain tumor treatment.

**Introduction**

According to the American Cancer Society[1], in 2018 nearly 24,000 patients will be diagnosed with brain or other nervous system cancer and about 17,000 patients will die of the disease. Gliomas represent about 25% of all primary brain tumors and about 80% of all malignant tumors of the central nervous system[2]. Over half of gliomas are glioblastoma multiforme (GBM), which is the most malignant primary brain tumor. GBMs are infiltrative and normally lack a clear margin making complete resection nearly impossible. Maximal resection of gliomas has been associated with improved prognosis[3,4], although invasion and the bounds of functional cortex often limit extensive removal. Currently, technology for extending the limits of the tumor resection relies on intraoperative image-guided surgical navigation platforms, intraoperative magnetic resonance imaging (MRI), and intraoperative ultrasound[5]. Wide-field fluorescence illumination through the operative microscope has been utilized more recently in an attempt identify the margins of infiltrating tumors[6].

Regardless of the means for identifying the tumor margin, examining tissue samples during surgery is paramount, especially for neurosurgery. Rapid intraoperative assessment of tumor tissue remains key for planning the treatment and for guiding the surgeon to areas of suspected tumor tissue during the operation, or planning adjunct intraoperative or post-operative therapy. The standard for the preliminary intraoperative histopathological interpretation is frozen section biopsy. However, the frozen section biopsy method has inherent complications such as sampling error, tissue freezing and cutting artifacts, lack of immediate pathologist interactivity with the surgeon, time spent for tissue delivery, processing, and analysis reporting back to the operating room[7,8].

Handheld (i.e., size of a pen), portable confocal laser endomicroscopy (CLE) is undergoing exploration in brain tumor surgery because of its ability to produce precise histopathological information of tissue with subcellular resolution *in vivo* in real time during tumor resection [8–13].

CLE is a fluorescence imaging technology that is used with a combination with fluorescent drugs or probes. While a wide range of fluorophores have been used for CLE in gastroenterology and other medical specialties, fluorophore options are limited for in vivo human brain use due to potential toxicity[8,9,11,14]. Fluorescent dyes currently approved for use *in vivo* in the human brain include fluorescein sodium (FNa), indocyanine green (ICG), and 5-aminolevulinic acid (5-ALA)[9,15,16]. Other fluorescent dyes such as acridine orange (AO), acriflavine (AF), cresyl violet, etc., can be used on human brain tissue *ex vivo*[12,17]. In neurosurgical oncology, CLE has been used to rapidly obtain optical cellular and cytoarchitectural information about tumor tissue as the resection progresses and to interrogate the resection cavity[12,13]. The details of system operation have been previously described in detail[8,12,13,18]. Briefly, the neurosurgeon may hold the CLE probe by the hand, fixate it with a flexible instrument holder in place, or may glide the probe across the tissue surface to obtain an "optical biopsy" with an image acquisition speed ranging between 0.8 and 20 frames per second dependent on operation of the particular CLE system. The surgeon may place the probe in a resting position at any time and proceed with the tumor resection, then take up the probe conveniently as desired. CLE imaging is believed to be potentially advantageous for appraisal of tumor margin regions or to examine suspected invasion into functional cortex near the final phases of tumor resection. The images display on a touchscreen monitor attached to the system. The neurosurgeon uses a foot pedal module to control depth of scanning and image acquisition. An assistant can also control the acquisition of images using a touchscreen. CLE images can be processed and presented as still images, digital video loops showing motion, or 3-dimensional digital imaging volumes. CLE is a promising technology with the strategy to optimize or maximally increase the resection of malignant infiltrating brain tumors and/or to increase the positive yield of tissue biopsy. CLE may be of especial value during surgery when interrogating tissue at the tumor border regions or within the surgical resection bed that may harbor remnant malignant or spreading tumor.

Cancer is the subject of intense investigation into how theranostics may improve care and survival. As oncology is continually refined in its quest to understand and treat malignant brain tumors, such as GBMs, with which it has had very little success, utilization of precision and personalized surgical techniques would seem to be a logical step forward, especially as tumor resection is usually the first definitive treatment step. Dealing with the abundance of images from CLE is not unlike sifting through a myriad of genes, proteins, or other structural or metabolic markers to find something of commonality or uniqueness in cancer that might indicate a potential treatment scheme or target. CLE data acquisition is vast, burdened with a near-overwhelming number of images, many of which appear not useful at first inspection, although they may have unrecognized informative image subregions or characteristics. Mathematical algorithms and computer-based technology may rapidly assist making decisions upon an incredible number of images, such as CLE produces, that has never been encountered in neurosurgery.

Critical success in theranostics relies on the analytical method. Finding meaning can be elusive, and what may seem at first meaningful may only be superficial or even a spurious result, thus the analytical methodology is critical. In this review we provide a detailed description of bioinformatical analysis methodology of CLE images that begins to assist the neurosurgeon and pathologist to rapidly connect on-the-fly intraoperative imaging, pathology, and surgical observation into a conclusionary system within the concept of theranostics. We describe methodology of deep convolutional neural networks (DCNNs) applied to CLE imaging focusing on neurosurgical application and review current modeling outcomes, elaborating and discussing studies aiming to suggest a more precise and tailored surgical approach and workflow for brain tumor surgery.

**Demanding imaging information load of confocal laser endomicroscopy**

Although the number of non-diagnostic CLE images has been shown to be high, the first diagnostic frames were acquired at an average after the 14th frame (about 17 sec) *in vivo*[12]. This is certainly faster than for an intraoperative frozen section biopsy preparation and diagnostic interpretation. Nevertheless, the high number of non-diagnostic images imposed a significant time requirement and image storage load for subsequent image reviews, leading us and other groups to employ deep learning algorithms and neural networks that could potentially sort out non-diagnostic frames, while retaining only the diagnostic ones[19,20]. Attempts to use advanced feature coding schemes to classify cellular CLE images of brain tumor samples stained *ex vivo* with AF have been reported[21]. Advantageously, acquired CLE images may be exchanged and translated for off-site digital histopathology review. However, large amounts of data may create an information overload that requires novel solutions for data CLE management and storage.

While CLE has obvious benefits of rapid on-the-fly digital imaging of tissue that can obviate long wait times for tissue interpretation and be quickly communicated between surgeons and pathologists, there are challenges to manage the amount of information provided. Current CLE systems can generate hundreds to thousands of images over the course of examination of the tumor or resection cavity which may take only a few minutes. It has been estimated that since CLE technology was put into use in 2011 for gastrointestinal diagnosis, over 100 million images have been created, with 30 million images created in the past year[22]. The number of images may become rapidly overwhelming for the neurosurgeon and neuropathologist when trying to review and select a diagnostic or meaningful image or group of images as the surgical inspection progresses. CLE is designed to be used in real time while the surgeon operates on the brain, but overcoming the barriers of image selection for diagnosis is a key component for making CLE a practical and advantageous technology for the neurosurgical operating room.

Other barriers for revealing underlying meaningful histology are motion and blood artifacts (especially) that are present in some of the CLE images, especially for CLE systems functioning in the blue laser range versus near-infrared[8,9,12,23,24]. In addition, the neuropathologist must begin to work in a world of fluorescence images showing shades of gray, black and white or artificial colorization, where before natural colored stains existed. The display of suboptimal nondiagnostic frames interferes with the selection of and focus upon diagnostic images by the neurosurgeon and pathologist throughout the surgery to make a correct intraoperative interpretation. A previous study of CLE in human brain tumor surgeries found that about half of the acquired images were interpreted as nonuseful (i.e., nondiagnostic) due to an inherent nature of the handheld microscopic probe with a narrow field of view that is subject to motion and blood artifacts or lack of discernible or characteristic features of the tissue itself[12]. These artifacts or inherent aspects of operation of the probe include unsteady hand movements, moving the probe while in imaging mode across the tissue surface, and irregularities of the tissue surface such as a tumor resection bed in the cortex that includes tissue crevices, surface irregularities, bleeding, movement of the cortex with arterial pressure and respiration, etc.

Thus, although imaging is acquired on-the-fly, an image discrimination system to optimally sift out and identify useful images would substantially improve the performance of the CLE. Manually filtering out the nondiagnostic images before making an intraoperative decision is challenging due to the large number of images acquired, the novel and frequently unfamiliar appearance of fluorescent stained tissue features compared to conventional histology. Interpretation of fluorescence CLE images for routine clinical pathology has only recently been trained and studied. Great variability among images from the same tumor type, and potential similarity between images from other tumor types for the untrained interpreter (Figure 1) make simple image filters and thresholding unreliable, thus requiring advanced computational methods.

**Theranostics and confocal laser endomicroscopy**

Investigations into cancer genetics have produced treatment pathways by the application of bioinformatics methods leading to the concept of theranostics. As advances in molecular science have enabled "fingerprinting" of individual tumor with genomic and proteomic profiling, personalized theranostic agents can be developed to target specific tumor microenvironment compartments[25]. Although theranostic imaging provides new opportunities for personalized cancer treatment through the interface of chemistry, molecular biology, and imaging, quantitative image analysis remains as one of its challenges[26]. More sophisticated image analysis methods are required to visualize and target every aspect of the tumor microenvironment in combination with molecular agents[27]. CLE technology potentially allows a more personalized, precise, or tailored approach to the surgical procedure to remove an invasive brain tumor because of its capability to image at cell resolution intraoperatively on-the-fly. Fluorescent stains or markers allow the imaging and potential targeting of cells – nearly we are at the surgery of the "cell" as specific stains or fluorescent markers are developed. Whether, this technology has practicality or yields survival benefit for malignant invasive brain tumors awaits the results of the first substantial *in vivo* explorations.

**Deep learning application in CLE brain tumor diagnostics**

Deep convolutional neural networks (DCNNs) are a subset of "deep learning" technology, a machine learning subfield that has achieved immense recognition in the field of medical image analysis. Advances in computer-aided detection (CADe) and diagnosis (CADx) systems in ultrasound, magnetic resonance imaging, and computed tomography (CT) have been reviewed previously[28]. There have been only a few studies that investigated deep learning application to enhance the diagnostic utility of CLE imaging in brain tumors. The utilization of deep learning approaches is mainly focused around three goals: diagnostic image detection, tumor classification, and feature localization originating from image segmentation. Here we provide an overview of the basics of a deep learning methodology applicable to CLE images, summarize current results of brain tumor CLE image analysis using DCNNs, and juxtapose these with related works in other cancers.

A deep convolutional neural network (DCNN) consists of several layers, each having multiple units called feature maps[29]. The first layer includes the input images that will be analyzed. To produce the second layer's feature maps, each pixel of layer 2 is connected to a local patch of pixels in layer 1 through a filter bank followed by an activation function, which is usually a rectified linear unit (ReLU) because of its fast-to-compute property compared to other functions[30]. Model parameters are learned by minimizing the loss (the error between model prediction and the ground truth) using an optimization algorithm in two steps: forward and backward propagations[31,32]. To adjust the weights of filter banks, after each iteration of forward propagating the network, the derivatives of the loss function with respect to different weights are calculated to update the weights in a backward propagation[32]. A pooling layer accumulates the features in a smaller region by replacing windows of a feature map with their maximum or the average value. By stacking several convolutional, pooling, and fully connected layers, a DCNN can learn a hierarchy of visual representations to recognize class-specific features in images[29]. Figure 2 shows an example network architecture and how the feature maps are calculated to perform diagnostic brain tumor image classification.

The validity of recommendations resulted from the DCNN analysis greatly depends upon the ground truth established by the expert professional. Unlike other conventional surgical tissue examination modalities like hematoxylin and eosin (H&E) stained histopathological slides, the CLE images are novel to neuropathologists and neurosurgeons. Since the beginning of CLE investigation in brain tumor surgery, the ground truth was established by surgical biopsy and subsequent standard histopathology analysis acquired from the same location as the CLE "optical biopsy" and correlating the features on CLE images to the histopathological sections. Neuropathologists and neurosurgeons at a few select centers are correlating CLE features to histopathology in order to establish an expertise in reading CLE images, and such investigations are ongoing[10,12,18]. The experience in CLE image interpretation is imperative for meaningful DCNN analysis. However, as described later, delving deeper into the DCNN analysis of the CLE while using ground truth established by the standard histopathology, results in identification of novel CLE features and allows many more images that may be termed suboptimal to in fact become useful. The improvement in workflow and diagnostics, and thus theranostics in CLE, will be dependent upon robust computer learning architecture.

*Tumor classification*

One of the first deep learning approaches for making a diagnosis of a brain tumor type based on the CLE images was a cascaded deep decision network (DDN), a type of DCNN[33]. A network was trained for classification of glioma and meningioma images using their previously proposed multi-stage DDN architecture[34] for developing the model. The training process was as follows: LeNet, a relatively shallow CNN architecture initially proposed by LeCun, et al.[35] for handwritten digit recognition, was trained on the training dataset until it produced descent classification results on validation images. Then, the images were divided into two categories: *easy images* (classified correctly by the model with high confidence) and *challenging images* (classified either wrongly or even correctly yet with a low confidence). The challenging images were passed to the next stage for retraining. In the second stage, a convolutional stage and two fully connected layers and a softmax layer were stacked to the previous network and trained, while freezing the previous layers' parameters. After training the second stage on the challenging images from stage 1, the same process was repeated (finding confidence threshold, filtering the easy images, passing the challenging images to next stage, stacking the new layers to the previous network) until the model fails to improve on the validation dataset. After removing uninformative images using image entropy, a dataset was created of about 14,000 GBM and 12,000 meningioma images. The final proposed DDN could classify the GBM images with 86% accuracy while outperforming other methods such as SVM classifier applied on manual feature extraction, pre-trained networks, and shallow CNNs. [34]

We have previously developed an architecture to classify CLE images from experimental brain gliomas into three classes: tumor tissue, injured brain cortex tissue (no tumor) and normal brain cortical tissue. [36] This study was undertaken to examine the ability of CLE image analysis to discriminate between tumor tissue, tissue subjected to the minor tissue trauma that surgical resection produces, and normal brain tissue. FNa may extravasate in the first two situations potentially causing surgeon confusion. This classification model was inspired by Inception, which is a DCNN for classifying generic images.[37] Due to the small size of our training dataset (663 diagnostic images selected from 1,130 images acquired), we used fine-tuning to train the model with a learning rate of 0.001. We used a nested left-out validation method to estimate the model performance on images from new biopsies. Images were divided into 3 data sets based on biopsy level: training (n=446), validation (n=217), and test set (n=40). Model performance increased to 88% when images were classified using 2 classes only (tumor tissue or non-tumor tissue) which was only slightly lower than the neuropathologists' mean accuracy (90%). The sensitivity and specificity of the model in discriminating a tumor region from non-tumor tissue were 78% and 100%, respectively. The Area Under the ROC Curve (AUC) value for tumor/non-tumor tissue classification was 93%. Subgroup analysis showed that the model could discriminate CLE images from tumor and injury with 85% accuracy (mean of accuracy for neuropathologists was 88%), 78% sensitivity and 100% specificity. We expect that performance of the model will be improved in terms of accuracy and speed by going forth from a small experimental data set to operation on large clinical data sets.

*Diagnostic image classification*

Entropy-based filtering is one of the simplest ways to filter out non-diagnostic CLE images. In a study by Kamen, et al.[21], CLE images obtained from brain tumors were classified automatically. An entropy-based approach was used to remove the noninformative images from their dataset and two common brain tumors (meningioma and glioma) were differentiated using bag of words and other sparse coding methods. However, entropy might not be an ideal method since many nondiagnostic images have nearly as high entropy as diagnostic ones, as shown in Figure 3. Due to the large number of CLE images produced during surgery, importance of data pruning, as has been shown in our previous study[36], and the incompetency of entropy method, we developed a deep learning model for reliable classification of images into diagnostic and non-diagnostic categories[20]. A blinded neuropathologist and 2 neurosurgeons proficient with CLE image interpretation individually annotated all the images in our dataset. For each patient, the histopathological features of corresponding CLE images and H&E-stained frozen and permanent sections were reviewed and the diagnostic value of each image was examined. When a CLE image revealed any clear identifiable histopathological feature, it was labeled as diagnostic; otherwise it was labeled as nondiagnostic[12]. Table 1 provides the composition of our dataset. We tested the developed diagnostic frame detection models on 4,171 CLE images chosen from various patients isolated during training.

Aiming to improve the diagnostic image classification method, we then developed 42 new models, which included 30 single, and 12 ensemble models using two network architectures, three training regimes, and two ensemble methods[20]. During training of each single model, different sections of the dataset were used to reflect the diversity of training data in the developed models' knowledge. We exercised various training regimes to investigate how "deep" the training should be for CNNs applied to a CLE image classification problem to produce optimal (i.e., diagnostically useful) results. Depending on which layers of the network are being learned through training, we had three regimes. In the deep training (DT) regime, the whole model parameters were initialized randomly (training from scratch) and were updated through training. In the shallow fine-tuning (SFT) regime, the whole model weights, except the last fully connected layer, were initialized with the corresponding values from the pretrained model and their values were fixed during training. The last fully connected layer was initialized randomly and was tuned during training. In the deep fine-tuning (DFT) regime, all model weights were initialized with the corresponding values from the pretrained model and were tuned with nonzero learning rates. Our cross validation showed SFT and DFT experiments required 10 times smaller initial learning rates (0.001) compared to the DT regime (0.01). We also used a dropout layer (ratio = 0.5) and L2 regularization ($\lambda = 0.005$).

For this interobserver study, we created a validation review dataset consisting of 540 images randomly chosen from the test dataset in the second review. Two new neurosurgeons reviewed the validation review dataset without having access to the corresponding H&E-stained slides and labeled them as diagnostic or nondiagnostic. The ensemble of DCNN models for detecting diagnostic CLE images achieved 85% agreement with the gold-standard defined by the trained expert with subsequent confirmation by another independent observer (Figure 4), without considering or comparison to the H&E slide images. In comparison, the two trained neurosurgeons achieved 75% and 67% agreement with the gold-standard using only CLE images. These results indicated that when only CLE images were provided, the model could detect the diagnostic CLE images with better agreement to the H&E-aided annotation. The example CLE images assessed with our diagnostic analysis model are presented on Figure 5. In order to compare the power of deep learning models with filtering approaches used in other related studies, we used entropy as a baseline[21,33]. Subsequent evaluation of our test dataset of CLE images suggested that DCNN-based diagnostic evaluation has a higher agreement with the ground truth compared to the entropy-based quality assessment (Table 2).

*Feature localization and image segmentation*

Most of the current object localization studies in medical imaging use supervised learning that requires an annotated dataset for the training process. Physicians need to review the images and mark the location of interesting areas for each image, thus making it a costly and time-consuming process. Weakly supervised localization (WSL) methods have been proposed in computer vision to localize features using a weaker annotation, i.e., image-level labels instead of pixel-level labels.

We have previously investigated feature localization on brain tumor CLE images.[20] Following the training and testing of the DCNN model for diagnostic image classification, 8 out of 384 reviewed colored neuron activation maps from the first layer of the model were selected for 4 diagnostic CLE images representative for glioma. Selected activation maps highlighted diagnostic tissue architecture patterns in warm colors. Particularly, selected maps emphasized regions of optimal image contrast, where hypercellular and abnormal nuclear features could be identified, and could serve as diagnostic features for image classification (Figure 5, bottom row). Additionally, a sliding window method was successfully applied to highlight diagnostic aggregates of abnormally large malignant glioma cells and atypically hypercellular areas[20] (Figure 5). Such feature localization from the hidden layers makes the interpretation of the model results more illustrative and objective, especially from a clinical point of view where diagnosis cannot be made without sufficient evidence. Additionally, model-based feature localization can be performed considerably faster than human inspection and interpretation.

In another study, we applied a state of the art WSL approach[38] to localize glioma tumor features in CLE images[39]. In this method, a global average pooling (GAP) layer was stacked to the convolutional layers of the network to create diagnostic feature maps. Representative localization and segmentation results are shown in Figures 6 and 7. A neurosurgeon with expertise in CLE imaging identified and highlighted the cellular areas in each CLE image (first column of both figures). By inserting the images and their labels (i.e., overall diagnostic quality: diagnostic and nondiagnostic) to the network, the model automatically learns the primary diagnostic features of gliomas (e.g., cellular areas). In Figure 6, the first column shows three CLE images along with the annotated diagnostic areas (red arrows) by a neurosurgeon, while the second column presents the diagnostic areas that the model highlighted with warm colors. The color bar near to each intensity map shows the relative diagnostic value for each color -- red marks the most diagnostic regions and blue marks the nondiagnostic regions. In Figure 7, after producing the diagnostic feature maps, each image was then segmented into diagnostic and nondiagnostic regions by thresholding (highlighted in green and purple); the recognized diagnostic regions correlated well with the neurosurgeon's annotation. This method has two potential benefits: 1) improvement of the efficiency of glioma CLE imaging by recognizing the present diagnostic features and guiding the surgeon in tumor resection; and, 2) further investigation of the detected diagnostic regions may extend the physician's perceptions about the glioma appearance and its phenotypes in CLE images.

**Deep learning-empowered CLE diagnostics in other cancers**

*Oral squamous cell carcinoma*

Oral squamous cell carcinoma (OSCC) is a common cancer affecting 1.3 million cases worldwide annually[40]. Because of the insufficient precision in current screening methods, most OSCC cases are unfortunately diagnosed at advanced stages leading to poor clinical outcome. CLE has allowed in vivo examination of OSCC which may lead to earlier and more effective therapeutic outcomes during examination[41]. In a study by Aubreville, et al.[42], a CNN was trained to classify normal and carcinogenic CLE image patches. A dataset of 11,000 CLE images was evenly distributed between the two classes. The images were acquired from 12 patients and images with artifact (motion, noise, mucus or blood) were excluded, leading to 7,894 good quality images. Consequently, each image was divided into 21 overlapping patches, all of which were labeled the same as the whole image. The artifact patches were removed from images and the remaining ones were normalized to have zero mean and unit standard deviation. Image rotation was used to augment the image dataset size.

LeNet was used to train the model for patch classification. This network has only two convolutional and one fully connected layers with drop out. The model combined the probability scores from each constituent patch as being carcinogenic to arrive at the final prediction for the whole image. The network was trained from scratch with initial learning rate of 0.001 and Adam optimizer to minimize the cross-entropy.

To compare the proposed method with conventional textural feature-based classification approaches, two feature extraction methods (Gray-Level Co-occurrence Matrix (GLCM) and Local Binary Patterns (LBP)) and a classification approach (Random Forest (RF)) methods were combined to discriminate images at two scales (1.0 x and 0.5 x). Furthermore, CNN transfer learning was explored by shallow fine-tuning the last fully connected layer of the pretrained Inception-v3 network[43], using the original dataset. For cross validation, a leave-one-patient-out cross validation was followed, meaning images were used from one patient for testing the model and the remaining cases for training the model.

Both the patch-based and whole image CNN approaches outperformed the textural feature extraction and classification methods. The proposed CNN method could differentiate the normal and carcinogenic CLE images with 88% accuracy, 87% sensitivity and 90% specificity when applied at 0.5x scale (the 1x scale produced suboptimal results). The shallow fine-tuned Inception-v3 model could also achieve 87% accuracy, 91% sensitivity and 84% specificity. The AUC values for the two methods (the proposed CNN and Inception-v3) were roughly similar (95%). The AUC values for feature extraction methods and RF classifier was significantly lower than CNN methods (RF-GLCM = 81%, RF-LBP = 89%). Interestingly, the trained model on OSCC CLE images was successfully applied for classification of CLE images from a different organ site, vocal cord squamous cell carcinoma.

In transfer learning with pretrained Inception-v3 the authors only modified the weights of the last layer, while keeping the previous layers parameters stationary. However, studies[20,44] have shown that deeply fine-tuning the pretrained networks may help the network adapt better to the new dataset by upgrading the feature extraction layers as well. However, shallow fine tuning only allows updating the classification layer, which might not be sufficient for optimal performance.

*Vocal cord cancer*

To differentiate between healthy and cancerous tissue of vocal cords, Vo, et al.[45] developed a Bag of Words (BoW) based on textural and CNN features using a dataset of 1,767 healthy and 2,657 carcinogenic images from five patients. Small patches with 105x105 pixel size were extracted and augmented (with rotation), leading to 374,972 patches. For the textural feature-based classification, each image was represented by the concatenation of all its constituent patch-driven feature descriptors. For the CNN features, a LeNet shallow CNN was trained on the patch dataset for a binary classification (with SGD optimizer; momentum = 0.9 and learning rate = 0.0005). To create the visual vocabulary, two feature encoding (Fisher vector[46] and Vector of Locally Aggregated Descriptors (VLAD)[47]) and two classification methods (SVM and RF) were tested for comparing their classification performance.

A Leave-One-Sequence-Out (LOSO) cross-validation was used to evaluate these methods. The CNN features combined with VLAD encoder and RF classifier achieved an accuracy of 82% and sensitivity of 82% on the test images that surpassed other approaches. However, despite its promising accuracy, the proposed multi-stage approach (patch creation, feature extraction, feature encoding, clustering and classification) is much more complicated than the current end-to-end DCNN architectures, which have all these procedures embedded in their stacked layers. However even with this approach, the CNN features could outperform textural features extracted manually[45].

*Lung cancer*

Gil, et al.[48] investigated visual patterns in bronchoscopic CLE images for discriminating benign and malignant lesions and aiding lung cancer diagnosis. A pretrained network developed by the Visual Geometry Group (VGG)[49] on a large generic image dataset was used for feature extraction and reduced the resulting feature vector dimension from 4,096 to 100, while preserving roughly 90% of the original feature vector energy for computational efficiency. Three different methods (k-Means, k-Nearest Neighbor (kNN), and their proposed topology-based approach) were applied on the feature codes to group images with similar features together and intrinsically discriminate images from benign and malignant tissue. Model predictive power was compared with the final diagnosis on 162 images from 12 cases (6 with malignant and 6 with benign lesions) and achieved 85% accuracy, 88% sensitivity, and 81% specificity.

Inter-observer studies were performed with three observers to compare the subjective visual assessment of images with the model performance (the observers were blinded to the final diagnosis). Interestingly, the three observers could make a correct diagnosis only for 60% of the selected CLE images (sensitivity: 73% for malignant and 36% for benign images) on average. In the second experiment, two observers made a final diagnosis after examining all the images from each case. The model was also supplied with all the images from each of the 12 cases and rendered a final decision for each case. While the model could differentiate malignant and benign cases with 100% accuracy (12/12), the two observers could confirmatively make the correct diagnosis only in 67% (8/12) of cases.

Although the observers' knowledge in the domain might have affected their performance, the objective results suggest that the bronchoscopic CLE images contain enough visual information for determining the malignancy of the tumor and the VGG network is an excellent candidate for extracting these discriminative features. The proposed topology-based clustering method could outperform common clustering and classification methods (K-means and kNN) in differentiating the two classes of images.

Despite its advantages, the proposed method had two major limitations. First, even though it can differentiate images from the two classes, it cannot predict the label for each cluster. The method can separate the images into two groups, but it is not able to give information about their labels. Second, it is unclear if there was independent development (for determining model parameters) and test datasets to avoid bias in model development.

*Gastrointestinal tract cancer*

Hong, et al.[50] proposed a CNN architecture for classifying CLE images from three subcategories of Barret's esophagus: intestinal metaplasia (IM), gastric metaplasia (GM), and neoplasia (NPL). The network was composed of four convolutional layers and two max-pooling and fully connected layers. The size of convolutional kernels was 3x3 and zero padding was also used. Stride of max-pooling was 2x2 which was applied in layers two and four. Fully connected layers followed the fourth convolutional layer, and each had 1,024 neurons. The output label was determined by a softmax layer which produced 3 probabilities for each subcategory.

The network was trained on the augmented CLE images of Barret's esophagus (155 IM, 26 GM and 55 NPL) for 15,000 iterations with the batch size of 20 images. Cross-entropy was used as a cost function in their experiment. The trained model was then tested on 26 independent images (17 IM, 4 GM and 5 NPL) for validation. The imbalance in size of different subcategories caused the model to observe more frequent instances of IM and NPL compared to GM during training. This created a bias in the model prediction which can be seen in the high accuracy for predicting IM and NPL instances (100% and 80%) and very low accuracy for predicting GM instances (0%). However, CLE is being used with increasing frequency for detecting pre-cancerous and cancerous lesions in the gastrointestinal (GI) tract. The highest numbers of CLE images have been acquired from the GI tract where such imaging technology has been approved for use clinically for a few years.

**Conclusions**

Precision, personalization, and improved therapeutics in medicine can only progress with improved technology, analysis, and logic. The science and philosophy of theranostics is the nexus of these. Several studies have emphasized the importance of theranostic imaging in personalized treatment of cancer[25–27]. Medical data acquired on patients has become more voluminous, and it will continue in such manner. The amount of data available and necessary for analysis has already eclipsed human capabilities. For example, the new technology of handheld surgical tools that can rapidly image at the cellular resolution on-the-fly produces more images than a pathologist can possibly examine. As CLE technology develops, there will not be one fluorophore, but multiple fluorophores applied directly to the tumor or administered to the patient varying from nonspecific to specifically identifying cell structures or processes used simultaneously and presented in a myriad of image combinations for greatly varying histopathology. Analytic methods for selection and interpretation of the CLE images is already being explored to be incorporated into CLE operating systems so that the unit display can differentiate tissue and label the image as well with near-on-the-fly capabilities. Computational hardware power and effective analytic model infrastructure are the only two limits. CLE systems and other related systems are being produced by several imaging technology companies and groups and are close to approval with European and American medical device regulatory agencies. However, it seems prudent given the enormous numbers of images already produced and those projected with adoption of such technology, that there is immediate exploration into such image analysis methods to allow the pathologist and neurosurgeon to make optimal decisions based on the CLE imaging, and to take advantage of the on-the-fly technology proposition.

Success or meaningful diagnostic and therapeutic indication in the burgeoning field of theranostics is only as good as the data incorporated and the methodology employed for analysis and to extract meaning, including its validation. In many cases relatively simple statistics have been used for analysis, while pattern recognition or neural network techniques may be used in more complicated scenarios. For images such as from CLE, the whole image may be important, or perhaps only certain subregions, or crucial data may lie in regions on cursory inspection deemed to be nonuseful, such as in areas of motion artifact. Complicating this situation are the overwhelming numbers of images yielded from the CLE application. Clinical decision environments currently require assistance to not only access and categorize the collection of images, but to also draw conclusions and inferences that have critical diagnostic and treatment consequences. A pathologist and neurosurgeon will not have time to inspect a thousand images per case, especially in the midst of CLE use intraoperatively. Therefore, a theranostics approach, i.e., the nexus of biological data, rapid informatics scrutiny and evaluation, and tailored human decision, must be employed as we venture into realms of ever increasing information in neurosurgery in search of personalization and precision, especially as we have encountered it first in the surgery and treatment of malignant invasive brain tumors. Additionally, pathologists and neurosurgeons will need to become versed in the methodology of the CLE decision making processes to have confidence in diagnostic labels and to base treatment decisions upon them, thus the reason for presenting details of analytical architectures in this review.

Two DCNN based approaches are reviewed in this paper: models that can automatically classify CLE images (classifications of images that are diagnostic/nondiagnostic[20], glioma/nonglioma[19], tumor/injury/normal[36]) and models that can localize histological features from diagnostic images using weakly supervised methods[39]. Manually annotated in-house datasets were used to train and test these approaches in most of the studies. For the tumor classification purpose, data pruning could enhance the results for both DCNN models and outperformed manual feature extraction and classification[21]. Fine-tuning and ensemble modeling could enhance the model performance in the diagnostic image classification. The ensemble effect was stronger in DT and DFT than SFT developed models.

Despite extensive research on CLE clinical application in neurosurgery[9–14,18], there have been few attempts in the automatic analysis of these images to enhance CLE clinical utility. Deep learning could be beneficial in filtering the nondiagnostic images with higher speed and reasonable accuracy compared to subjective assessment.[19] Our inter-rater agreement evaluation[20] showed that the proposed model could achieve promising agreement with the gold-standard defined by a majority assessment by neurosurgeon reviewers. Overall, results suggest that DCNN-based diagnostic evaluation has a higher agreement with the ground truth than the entropy-based quality assessment used in other studies[21,33]. Furthermore, such methods suggest that semantic histological features may be highlighted in CLE images as confirmed by a neurosurgeon reviewer. This shows that the DCNN structure could learn semantic concepts like tumor type or diagnostic value of CLE images through different levels of feature representation. Early results show that WSL-based glioma feature localization was able to precisely mark the cells in the images. DCNNs are also much faster than handcrafted methods at deployment phase. Our deeply trained models could classify about 40 new images in a second, while the handcrafted method takes 5.4 seconds to process single image[21].

Other confocal imaging techniques may be aided by such deep learning models. Confocal reflectance microscopy (CRM) has been studied[51,52] for rapid, fluorophore-free evaluation of brain biopsy specimen *ex vivo*. CRM allows preserving the biopsy tissue for future permanent analysis, immunohistochemical studies, and molecular studies. Proposed DCNN classification and localization approaches are well-suited for analysis and interpretation of CRM images as well as CLE. Further studies on application of DCNN on CRM images are needed to further validate their utility for intraoperative diagnosis.

Continued use of unsupervised image segmentation methods to detect meaningful histological features from confocal brain tumor images will likely allow for more rapid and detailed diagnosis. With the large rate of images produced, a technology-free and unassisted approach to analyze the CLE images would impede the exploitation of maximal pathological information during surgery. Accessible databases of CLE images would allow various image analysis methods to be tried on large numbers of images. Such image collection strategies are part of the platform of the relatively new International Society for Endomicroscopy. DCNNs can enhance extraction and recognition of CLE diagnostic features that may be integrated into the standard brain tumor classification protocols similarly to the current research flow in the whole-slide digital image analysis for personalized cancer care[53,54]. This may refine current diagnostic criteria and potentially aid the discovery of novel related features. With such technology, neurosurgery truly enters the realm of theranostics in the operating room itself—we are on the verge of highly tailored and precise surgery at the cellular level. Such an approach is critical for neurosurgery because surgery and treatment for an invasive brain tumor frequently deals with spread into eloquent cortex – the areas that make us "human." In fact, even before entering the operating room the neurosurgeon can begin to discuss strategy with the patient if tumor is located or not located in eloquent cortex based on a CLE "optical biopsy". Thus, theranostics also involves treatment strategies and decisions of when to "stop", especially true when the CLE system intraoperatively reveals cells invading for example primary motor or language cortex. With analytical pathologists uniting different clinical and morphological information for an integrated diagnosis, such a computer-aided CLE analysis workflow would improve imaging (diagnostics) and achieve maximal, more precise removal of tumor mass (therapy) as the initial treatment goals toward greater precision, personalization and success in the surgery and treatment of malignant invasive brain tumors (theranostics).

**Acknowledgments**

We thank the Barrow Neurological Foundation; the Newsome Chair in Neurosurgery Research to MCP for funding. EB acknowledges stipend support SP-2240.2018.4. Part of this study involving Confocal Laser Endomicroscopy was supported by a grant from Carl Zeiss AG, Oberkochen, Germany. The grantor did not have any contribution or effect on study design, data collection, analysis or paper preparation.

# REFERENCES

1. American Cancer Society. Cancer Facts and Statistics. (2018). Available at: https://cancerstatisticscenter.cancer.org/. (Accessed: 29th May 2018)
2. Ostrom, Q. T. *et al.* CBTRUS Statistical Report: Primary Brain and Central Nervous System Tumors Diagnosed in the United States in 2008-2012. *Neuro. Oncol.* **17,** iv1-iv62 (2015).
3. Almeida, J. P., Chaichana, K. L., Rincon-Torroella, J. & Quinones-Hinojosa, A. The Value of Extent of Resection of Glioblastomas: Clinical Evidence and Current Approach. *Current Neurology and Neuroscience Reports* **15,** (2015).
4. Sanai, N., Polley, M.-Y., McDermott, M. W., Parsa, A. T. & Berger, M. S. An extent of resection threshold for newly diagnosed glioblastomas: clinical article. *J. Neurosurg.* **115,** 3–8 (2011).
5. Sanai, N. & Berger, M. S. Surgical oncology for gliomas: the state of the art. *Nat. Rev. Clin. Oncol.* **15,** 112 (2018).
6. Maugeri, R. *et al.* With a Little Help from My Friends: The Role of Intraoperative Fluorescent Dyes in the Surgical Management of High-Grade Gliomas. *Brain Sci.* **8,** 31 (2018).
7. Tofte, K., Berger, C., Torp, S. H. & Solheim, O. The diagnostic properties of frozen sections in suspected intracranial tumors: A study of 578 consecutive cases. *Surg. Neurol. Int.* **5,** (2014).
8. Martirosyan, N. L. *et al.* Potential application of a handheld confocal endomicroscope imaging system using a variety of fluorophores in experimental gliomas and normal brain. *Neurosurg. Focus* **36,** E16 (2014).
9. Belykh, E. *et al.* Intraoperative fluorescence imaging for personalized brain tumor resection: Current state and future directions. *Front. Surg.* **3,** (2016).
10. Charalampaki, P. *et al.* Confocal Laser Endomicroscopy for Real-time Histomorphological Diagnosis: Our Clinical Experience With 150 Brain and Spinal Tumor Cases. *Neurosurgery* **62,** 171–176 (2015).
11. Foersch, S. *et al.* Confocal laser endomicroscopy for diagnosis and histomorphologic imaging of brain tumors in vivo. *PLoS One* **7,** e41760 (2012).
12. Martirosyan, N. L. *et al.* Prospective evaluation of the utility of intraoperative confocal

laser endomicroscopy in patients with brain neoplasms using fluorescein sodium: experience with 74 cases. *Neurosurg. Focus* **40,** E11 (2016).

13. Sanai, N. *et al.* Intraoperative confocal microscopy for brain tumors: a feasibility analysis in humans. *Neurosurgery* **68,** ons282--ons290 (2011).
14. Zehri, A. *et al.* Neurosurgical confocal endomicroscopy: A review of contrast agents, confocal systems, and future imaging modalities. *Surg. Neurol. Int.* **5,** 60 (2014).
15. Mooney, M. A., Zehri, A. H., Georges, J. F. & Nakaji, P. Laser scanning confocal endomicroscopy in the neurosurgical operating room: a review and discussion of future applications. *Neurosurg. Focus* **36,** E9 (2014).
16. Liu, J. T. C., Meza, D. & Sanai, N. Trends in fluorescence image-guided surgery for gliomas. *Neurosurgery* **75,** 61–71 (2014).
17. Martirosyan, N. L. *et al.* Confocal scanning microscopy provides rapid, detailed intraoperative histological assessment of brain neoplasms: Experience with 106 cases. *Clin. Neurol. Neurosurg.* **169,** 21–28 (2018).
18. Eschbacher, J. *et al.* In vivo intraoperative confocal microscopy for real-time histopathological imaging of brain tumors: Clinical article. *J. Neurosurg.* **116,** 854–860 (2012).
19. Izadyyazdanabadi, M. *et al.* Improving utility of brain tumor confocal laser endomicroscopy: Objective value assessment and diagnostic frame detection with convolutional neural networks. in *Progress in Biomedical Optics and Imaging - Proceedings of SPIE* **10134,** (2017).
20. Izadyyazdanabadi, M. *et al.* Convolutional Neural Networks: Ensemble Modeling, Fine-Tuning and Unsupervised Semantic Localization for Neurosurgical CLE Images. *J. Vis. Commun. Image Represent.* **54,** 10–20 (2018).
21. Kamen, A. *et al.* Automatic Tissue Differentiation Based on Confocal Endomicroscopic Images for Intraoperative Guidance in Neurosurgery. *Biomed Res. Int.* **2016,** (2016).
22. Loiseau, S. Presentation at International Society for Endomicroscopy, Paris, France. (2017).
23. Sankar, T. *et al.* Miniaturized handheld confocal microscopy for neurosurgery: Results in an experimental glioblastoma model. *Neurosurgery* **66,** 410–417 (2010).
24. Martirosyan, N. L. *et al.* Use of in vivo near-infrared laser confocal endomicroscopy with

indocyanine green to detect the boundary of infiltrative tumor. *J. Neurosurg.* **115,** 1131–1138 (2011).

25. Stasinopoulos, I. *et al.* Exploiting the tumor microenvironment for theranostic imaging. *NMR in Biomedicine* **24,** 636–647 (2011).

26. Penet, M. F., Chen, Z., Kakkad, S., Pomper, M. G. & Bhujwalla, Z. M. Theranostic imaging of cancer. *Eur. J. Radiol.* **81,** (2012).

27. Penet, M.-F., Krishnamachary, B., Chen, Z., Jin, J. & Bhujwalla, Z. M. Molecular imaging of the tumor microenvironment for precision medicine and theranostics. *Adv. Cancer Res.* **124,** 235–56 (2014).

28. Greenspan, H., van Ginneken, B. & Summers, R. M. Guest editorial deep learning in medical imaging: Overview and future promise of an exciting new technique. *IEEE Trans. Med. Imaging* **35,** 1153–1159 (2016).

29. LeCun, Y., Bengio, Y. & Hinton, G. Deep learning. *Nature* **521,** 436–444 (2015).

30. Glorot, X., Bordes, A. & Bengio, Y. Deep sparse rectifier neural networks. *AISTATS '11 Proc. 14th Int. Conf. Artif. Intell. Stat.* **15,** 315–323 (2011).

31. Rumelhart, D. E., Hinton, G. E. & Williams, R. J. Learning representations by back-propagating errors. *Nature* **323,** 533–536 (1986).

32. Le Cun Jackel, B. Boser, J. S. Denker, D. Henderson, R. E. Howard, W. Hubbard, L. D., Cun, B. Le, Denker, J. & Henderson, D. Handwritten Digit Recognition with a Back-Propagation Network. *Adv. Neural Inf. Process. Syst.* 396–404 (1990). doi:10.1111/dsu.12130

33. N. Murthy, V. *et al.* Cascaded deep decision networks for classification of endoscopic images. in *Medical Imaging 2017: Image Processing* (eds. Styner, M. A. & Angelini, E. D.) **10133,** 101332B (2017).

34. Murthy, V. N., Singh, V., Chen, T., Manmatha, R. & Comaniciu, D. Deep Decision Network for Multi-class Image Classification. in *2016 IEEE Conference on Computer Vision and Pattern Recognition (CVPR)* 2240–2248 (2016). doi:10.1109/CVPR.2016.246

35. LeCun, Y. *et al.* Learning algorithms for classification: A comparison on handwritten digit recognition. in *Neural networks: the statistical mechanics perspective* 261–276 (1995).

36. Belykh, E. *et al.* Diagnostic accuracy of the confocal laser endomicroscope for in vivo differentiation between normal and tumor tissue during fluorescein-guided glioma resection: Laboratory investigation. *World Neurosurg.* **In press,** (2018).

37. Szegedy, C. *et al.* Going deeper with convolutions. in *Proceedings of the IEEE Conference on Computer Vision and Pattern Recognition* 1–9 (2015). doi:10.1109/CVPR.2015.7298594

38. Zhou, B., Khosla, A., Lapedriza, A., Oliva, A. & Torralba, A. Learning deep features for discriminative localization. in *Proceedings of the IEEE Conference on Computer Vision and Pattern Recognition* 2921–2929 (2016). doi:10.1109/CVPR.2016.319

39. Izadyyazdanabadi, M. *et al.* Weakly-Supervised Learning-Based Feature Localization in Confocal Laser Endomicroscopy Glioma Images. *arXiv Prepr. arXiv1804.09428* (2018).

40. Ferlay, J. *et al.* Cancer incidence and mortality worldwide: Sources, methods and major patterns in GLOBOCAN 2012. *Int. J. Cancer* **136,** E359–E386 (2015).

41. Thong, P. S.-P. *et al.* Laser confocal endomicroscopy as a novel technique for fluorescence diagnostic imaging of the oral cavity. *J. Biomed. Opt.* **12,** 14007 (2007).

42. Aubreville, M. *et al.* Automatic Classification of Cancerous Tissue in Laserendomicroscopy Images of the Oral Cavity using Deep Learning. *Sci. Rep.* **7,** (2017).

43. Szegedy, C., Vanhoucke, V., Ioffe, S., Shlens, J. & Wojna, Z. Rethinking the Inception Architecture for Computer Vision. in *2016 IEEE Conference on Computer Vision and Pattern Recognition (CVPR)* 2818–2826 (IEEE, 2016). doi:10.1109/CVPR.2016.308

44. Tajbakhsh, N. *et al.* Convolutional neural networks for medical image analysis: full training or fine tuning? *IEEE Trans. Med. Imaging* **35,** 1299–1312 (2016).

45. Vo, K., Jaremenko, C., Bohr, C., Neumann, H. & Maier, A. Automatic Classification and Pathological Staging of Confocal Laser Endomicroscopic Images of the Vocal Cords. in *Bildverarbeitung für die Medizin 2017* 312–317 (Springer, 2017).

46. Sánchez, J., Perronnin, F., Mensink, T. & Verbeek, J. Image classification with the fisher vector: Theory and practice. *Int. J. Comput. Vis.* **105,** 222–245 (2013).

47. Jégou, H., Douze, M., Schmid, C. & Pérez, P. Aggregating local descriptors into a compact image representation. in *Proceedings of the IEEE Computer Society Conference on Computer Vision and Pattern Recognition* 3304–3311 (2010). doi:10.1109/CVPR.2010.5540039

48. Gil, D. et al. Classification of Confocal Endomicroscopy Patterns for Diagnosis of Lung Cancer. in *Cardoso M. et al. (eds) Computer Assisted and Robotic Endoscopy and*

*Clinical Image-Based Procedures. CARE 2017, CLIP 2017. Lecture Notes in Computer Science* **10550,** 151–159 (2017).

49. Chatfield, K., Simonyan, K., Vedaldi, A. & Zisserman, A. Return of the Devil in the Details: Delving Deep into Convolutional Nets. *BMVC* 1–11 (2014). doi:10.5244/C.28.6

50. Hong, J., Park, B. & Park, H. Convolutional neural network classifier for distinguishing Barrett's esophagus and neoplasia endomicroscopy images. in *2017 39th Annual International Conference of the IEEE Engineering in Medicine and Biology Society (EMBC)* 2892–2895 (IEEE, 2017). doi:10.1109/EMBC.2017.8037461

51. Eschbacher, J. M. *et al.* Immediate Label-Free Ex Vivo Evaluation of Human Brain Tumor Biopsies With Confocal Reflectance Microscopy. *J. Neuropathol. Exp. Neurol.* **76,** 1008–1022 (2017).

52. Mooney, M. A. *et al.* Immediate ex-vivo diagnosis of pituitary adenomas using confocal reflectance microscopy: a proof-of-principle study. *J. Neurosurg.* **128,** 1072–1075 (2018).

53. Madabhushi, A. & Lee, G. Image analysis and machine learning in digital pathology: Challenges and opportunities. *Medical Image Analysis* **33,** 170–175 (2016).

54. Djuric, U., Zadeh, G., Aldape, K. & Diamandis, P. Precision histology: how deep learning is poised to revitalize histomorphology for personalized cancer care. *npj Precis. Oncol.* **1,** 22 (2017).

55. Louis, D. N. *et al.* The 2007 WHO classification of tumours of the central nervous system. *Acta Neuropathologica* **114,** 97–109 (2007).

56. Krizhevsky, A., Sutskever, I. & Hinton, G. E. Imagenet classification with deep convolutional neural networks. in *Advances in neural information processing systems* 1097–1105 (2012).

**FIGURE CAPTIONS**

**Figure 1.** Representative Confocal Laser Endomicroscopy (CLE) images from glioma and meningioma acquired with Optiscan 5.1, Optiscan Pty., Ltd. (a) anaplastic oligodendroglioma, (b) recurrent astrocytoma, (c) glioblastoma multiforme (GBM), (d) fibrous meningioma (grade I), (e) chordoid meningioma (grade II), (f) atypical meningioma (grade II). Field of view = 475 × 475µm, resolution = 1024 × 1024 pixels, bar = 100 µm. (Glioblastomas are a brain malignancy of astrocytic cell origin, show wild pleomorphism, proliferation of abnormal tumor-associated vasculature, necrosis, and vast brain invasion. Meningiomas arise from meningothelial cells and are usually attached to the dura. Although of a common origin, meningiomas have histological pattern subtypes and more aggressive types show atypical or anaplastic features. They do not display malignant brain infiltration.[55])

**Figure 2.** Deep convolutional neural network (DCNN) architecture. A schematic diagram of AlexNet[56], a famous DCNN architecture that was trained on CLE images for diagnostic classification by Izadyyazdanabadi et al.[20], is shown in (a). Different feature maps of the first convolutional layer (color images) were calculated by convolving different filters (red squares) with the corresponding regions of the input image (illustrated in (b)).

**Figure 3.** Entropy of diagnostic (orange) and nondiagnostic (blue) images. The overlap between the entropy of diagnostic and nondiagnostic CLE frames limits its feasibility for precise discrimination between the two classes[19].

**Figure 4.** A schematic diagram of interobserver study. Gold standard was defined using the initial review and one of the secondary raters. The agreement of the ensemble model as well as the other rater with the gold standard is calculated[20].

**Figure 5.** Unsupervised semantic localization of the CLE histopathological features[20]. First row displays the original CLE images, along with the probability of each image being diagnostic (D) and nondiagnostic (ND), estimated by the model. Red arrows mark the cellular regions recognized by a neurosurgeon. Second row shows the corresponding activation of neurons from the first layer (conv1, neuron 24) (shallow features learned by the model); it highlights some of the cellular areas (in warm colors) present in the images which were identified as diagnostic regions by the neurosurgeon reviewer. The color bars show the relative diagnostic value for each color: red marks the most diagnostic regions (1.0) and blue marks the nondiagnostic regions (0.0). Field of view = 475 × 475µm, resolution = 1024 × 1024 pixels, bar = 100 µm.

**Figure 6.** Histological glioma feature localization with a weakly supervised approach: global average pooling (GAP)[39]. Left column shows CLE images from glioma cases ((a, c) recurrent infiltrating astrocytoma (e) oligodendroglioma). Red arrows mark the cellular regions recognized by a neurosurgeon. Second column shows the important regions detected with the model (highlighted in warm colors). The color bars show the relative diagnostic value for each color: red marks the most diagnostic regions (1.0) and blue marks the nondiagnostic regions (0.0). Field of view = 475 × 475µm, resolution = 1024 × 1024 pixels, bar = 100 µm.

**Figure 7.** Diagnostic image segmentation in CLE images of gliomas with GAP approach[39]. First column (a, d, e) shows the original diagnostic images from glioma cases ((a) recurrent GBM, (d) recurrent infiltrating astrocytoma, (e) anaplastic oligodendroglioma). Red arrows mark the cellular regions recognized by a neurosurgeon. Second column shows the segmented key features highlighted in purple. Field of view = 475 × 475µm, resolution = 1024 × 1024 pixels, bar = 100 µm.

**TABLES**

**Table 1.** The composition of our dataset in the diagnostic image classification[20]. Number of patients and images used for model development and testing is provided.

|  | Development | Test | Total |
|---|---|---|---|
| **No. of Patients** | 59 | 15 | 74 |
| **No. of Images** | 16,366 | 4,171 | 20,537 |
| No. of Diagnostic Images | 8,023 | 2,071 | 10,094 |
| No. of Nondiagnostic Images | 8,343 | 2,100 | 10,443 |

**Table 2.** DCNN and entropy-based performance in diagnostic image classification[20]. DCNN methods showed higher agreement with the neurosurgeons' evaluation. DCNN2[37] has a deeper architecture with fewer parameters than DCNN1[56].

| Model | Accuracy (%) | AUC |
|---|---|---|
| DCNN 1 | 78.8 | 0.87 |
| DCNN 2 | 81.8 | 0.89 |
| Entropy-based | 57.20 | 0.71 |